# Enhancing Keyphrase Extraction from Academic Articles Using Section Structure Information

Chengzhi Zhang [a, *], Xinyi Yan [a], Lei Zhao [a], Yingyi Zhang[b]

[a] Department of Information Management, Nanjing University of Science and Technology, Nanjing, China

[b] Department of Archives and E-government, Soochow University, Suzhou, China

**Abstract**: The exponential increase in academic papers has significantly increased the time required for researchers to access relevant literature. Keyphrase Extraction (KPE) offers a solution to this situation by enabling researchers to efficiently retrieve relevant literature. The current study on KPE from academic articles aims to improve the performance of extraction models through innovative approaches using *Title* and *Abstract* as input corpora. However, the semantic richness of keywords is significantly constrained by the length of the abstract. While full-text-based KPE can address this issue, it simultaneously introduces noise, which significantly diminishes KPE performance. To address this issue, this paper utilized the structural features and section texts obtained from the section structure information of academic articles to extract keyphrase from academic papers. The approach consists of two main parts: (1) exploring the effect of seven structural features on KPE models, and (2) integrating the extraction results from all section texts used as input corpora for KPE models via a keyphrase integration algorithm to obtain the keyphrase integration result. Furthermore, this paper also examined the effect of the classification quality of section structure on the KPE performance. The results show that incorporating structural features improves KPE performance, though different features have varying effects on model efficacy. The keyphrase integration approach yields the best performance, and the classification quality of section structure can affect KPE performance. These findings indicate that using the section structure information of academic articles contributes to effective KPE from academic articles. The code and dataset supporting this study are available at https://github.com/yan-xinyi/SSB_KPE.

[*] Corresponding author: Chengzhi Zhang (zhangcz@njust.edu.cn).

## 1. Introduction

With the continuous development of scientific research, the volume of academic articles has surged, resulting in an overwhelming amount of information (Bornmann & Mutz, 2015). This rapid growth presents a challenge for researchers seeking to access. Keyphrases, considered as the smallest units representing an article's topic (Nasar et al., 2019), play a crucial role in facilitating quick access to articles and addressing the issue of handling large volumes of literature. Keyphrase extraction (KPE) is widely applied in fields such as information retrieval (Chien, 1999), text classification (Hassaine et al., 2015), and text summarization (Liu et al., 2022), and remains a significant research area in natural language processing (NLP).

Many studies have advanced the performance of KPE models by utilizing the title and abstract as the input corpus, due to the complexity of dealing with longer sequences (Kontoulis et al., 2021). However, relying solely on abstracts can lead to incomplete information, loss of context, and semantic ambiguity. To surmount these challenges, scholars have gradually shifted their focus to full-text based KPE, leveraging the structural information of the entire text (such as paragraphs, sentences, headings, citations, etc.) to enhance KPE efficacy (Feng et al., 2011; Zhang et al., 2022; Ajallouda et al., 2022). However, full-text information introduces noise and context dependency, which would reduce the effectiveness of KPE. Hence, researchers have begun exploring the use of section structure information to enhance KPE, because academic articles are typically organized into structured sections, and identifying these sections can effectively filter out noise (Zhou & Li, 2020). Currently, despite significant advancements in section based KPE, challenges such as the diversity of section characteristics, the issue of section weighting, and data scarcity continue to hinder progress in this area. Many studies overlook the rich keyphrases in crucial sections like "Methods" and "Results," leading to incomplete or suboptimal results (Nguyen & Luong, 2010). To tackle this issue, there has been a growing focus on exploring the section structure of academic literature. Notably, Bao et al. (2024) introduced a standardized framework for structural function recognition, which facilitates a detailed analysis of discourse structures in scholarly articles. This approach offers fresh perspectives and innovative methodologies for advancing related research. Therefore, this paper proposes a keyphrase extraction framework that integrates both section structure features and section texts. It dynamically adjusts the predicted keyphrase list based on the importance of candidate terms in each section, aiming to obtain a high-quality, semantically rich set of keyphrases.

Our contributions are the following three folds:

Firstly, our research employs section structure features and textual content to enhance KPE, achieving significant improvements and laying a solid foundation for future studies.

Secondly, we constructed a section-annotated KPE corpus containing 5316 academic papers, offering crucial data support for research on section structure and KPE, and establishing a benchmark for related studies in other languages

Thirdly, we examine the impact of section structure classification quality on KPE. The results demonstrate that corpora with clear section structure yield superior KPE performance, while the KPE performance of corpora with vague section structure is not necessarily inferior.

All the data and source code of this paper are freely available at the GitHub website: https://github.com/yan-xinyi/SSB_KPE.

## 2. Related work

This paper leverages section structure information within academic articles for the KPE task. Therefore, the following two sections review the related work on section structure classification and automatic KPE for academic articles, respectively.

### 2.1 Automatic KPE of academic articles

Automatic KPE involves extracting phrases from an article that best encapsulate its content using machine learning methods. There are two primary approaches: the unsupervised approach, which does not require any training corpus, and the supervised approach, which necessitates training and testing samples.

**Table 1.** Summary of representative studies of unsupervised keyphrase extraction approaches

| Categories of methods | Authors | Methods | Main findings |
|---|---|---|---|
| Statistical feature-based methods | *Hassani et al., 2022* | *LVTIA* | The study developed an algorithm for keyphrase extraction from video lectures, demonstrating superior performance compared to other algorithms across various datasets in both English and Persian. |
| | *Goz & Mutlu, 2023* | *SkyWords* | This algorithm significantly decreased the number of candidate keywords and attained the highest mean average precision score across all datasets. |
| | *Chen, 2024* | *TF*IDF* | The study found that an extended Term Frequency-Inverse Document Frequency (TF-IDF) method effectively addresses the limitations of Dunning's Log-Likelihood Test (LLT) in keyword extraction. |
| Graph-based methods | *Li et al., 2017* | *TextRank* | It significantly improves weakly supervised semantic segmentation by progressively refining pixel-level annotations, achieving state-of-the-art performance with 64.7%. |
| | *Bellaachia & Al-Dhelaan, 2012* | *NE Rank* | This paper introduces NE-Rank, a new keyword ranking method, and shows significant improvements in keyphrase extraction from tweets, achieving up to 39% improvement with NE-Rank and 20% improvement using hashtags.. |
| | *Vega-Oliveros et* | *Multi-Centrality* | The study optimally combines various centrality measures for keyword extraction in documents, demonstrating that MCI significantly outperforms individual centrality |
| | *al., 2019* | *Index (MCI)* | methods and other clustering algorithms. |
| Topic-based approaches | *Hyung et al., 2017* | *LDA* | This study integrates context-relevant music descriptors with keyword extraction, demonstrating the potential of this approach in semantic music search and retrieval. |
| | *Chin et al., 2019* | *TOTEM* | TOTEM leverages LDA to generate the topic labels as well as the topic summaries associated with them. Evaluation of the generated topic summaries suggest reasonably high consistency and effectiveness of summaries. |
| | *Rodrigues et al., 2021* | *LDA, SPARK* | The research examines Twitter trends utilizing machine learning and big data techniques, attaining an 83% accuracy rate through Jaccard similarity and enhancing real-time analysis speed with Apache SPARK. |
| Embedding-based approaches | *Bennani-Smires, 2018* | *EmbedRank* | EmbedRank is a novel unsupervised method for single-document keyphrase extraction that leverages sentence embeddings to achieve superior performance of KPE over graph-based approaches. |
| | *Sharma & Li, 2019* | *keyBERT* | *keyBERT* introduces a novel self-supervised end-to-end deep learning approach for KPE, leveraging contextual and semantic features to achieve superior performance compared to state-of-the-art methods |
| | *Patel & Caragea, 2021* | *KPRank* | As an unsupervised graph-based algorithm, KPRank integrates positional information and contextual word embeddings into a biased PageRank, achieving superior performance on KE across five benchmark datasets compared to previous methods and strong baselines. |

Referring to the Table 1, the unsupervised KPE approach predominantly relies on intrinsic document features within the text. Its core idea is to selects top N phrases with the highest weights as keyphrases. The most widely employed weight calculation methods include TF*IDF (Salton & Buckley, 1988) and TextRank (Mihalcea & Tarau, 2004). Unsupervised KPE's improvement studies can be categorized into statistical feature-based approaches (Goz & Mutlu, 2023), graph-based approaches (Huang et al., 2020), topic-based approaches (Hyung et al., 2017) and semantic-based approaches. Overall, statistical unsupervised methods are simple and efficient but may miss context; graph-based methods capture relationships well but can be complex; topic-based methods provide rich semantic information but require extensive computational resources. Compared to these methods, embedding-based KPE, such as KPRank, leverages the contextual richness of pre-trained word embeddings, allowing it to better capture semantic nuances and outperform traditional methods on benchmark datasets. However, embedding-based KPE often require significant computational power and may depend on the quality of pre-trained embeddings, making them less suitable for resource-constrained environments.

**Table 2.** Summary of representative studies of supervised keyphrase extraction approaches

| **Categories of methods** | **Authors** | **Methods** | **Main findings** |
|---|---|---|---|
| ML-based methods | *Wang et al., 2018* | *RF* | With the help of feature extraction discretization and classification, the prediction accuracy has been improved to a large extent. |
| | *Guleria et al., 2021* | *SVMs* | The work addresses keyphrase extraction by proposing a supervised machine learning method using SVM, resulting in significant improvements compared to established methods. |
| | *Lu & Chow, 2021* | *CRF* | CRF was utilized to encode segment-level features and model keyphrase duration, showing improved decoding performance across varied datasets. |
| DL-based methods | *Ray Chowdhury et al., 2019* | *LSTM* | The model shows enhanced performance on both general and disaster-specific Twitter datasets. |
| | Alzaidy et al., 2019 | *BiLSTM-CRF* | Through introducing a model that integrates CRF with BiLSTM, the results shown significant performance improvements over strong baselines on academic document datasets. |
| | Yan et al., 2024 | *Multiple models* | Tests on tweet datasets demonstrate the enhancement of cognitive signals generated during human reading on keyphrase extraction from Microblogs. |
| PLM-based methods | *Xiong et al., 2019* | *BLING-KPE* | The findings from OpenKP validate the efficacy of BLINGKPE, while zero-shot evaluations on DUC2001 highlight the enhanced generalization capability of training on open-domain data as opposed to domain-specific data. |
| | *Wang et al., 2020* | *SMART-KPE-full* | This approach, which integrates effective strategy induction and selection, surpasses state-of-the-art models for the keyphrase extraction (KPE) task. |
| | *Song et al., 2021* | *KIEMP* | Results on six benchmark datasets demonstrate that KIEMP surpasses most existing state-of-the-art keyphrase extraction methods. |

In contrast, the supervised KPE methods utilizes annotated corpora to train the model which is capable of predicting keyphrases in new articles. As can be seen from Table 2, Witten et al. (1999) employed a traditional machine method, naive Bayes algorithm, to classify candidate phrases and extract keyphrases from articles. Deep learning approaches are also popular in supervised KPE, Al-Zaidy et al. (2019) proposed a BiLSTM-CRF model, which outperformed previous methods in extracting keyphrases from articles. Several recent studies have also enhanced KPE performance by optimizing the input corpus, such as using automatic summarization (Kontoulis et al., 2021), incorporating reference information (Zhang et al., 2022). With the development of Pre-train Language Models (PLMs) and Large Language Models (LLMs), related KPE researches have also increased sharply. The current state-of-the-art PLM-based PKE models include SMART-KPE-full (Wang et al., 2020), RoBERTa-JointKPE , and KIEMP (Song et al., 2021). When it comes to LLMs with billion-level parameters, research based on T5 (Pęzik et al., 2022)and ChatGPT (Wei et al., 2023) is considered most representative.

Existing research suggests that supervised KPE methods necessitate the development of extensive training corpora to achieve optimal performance. Conversely, unsupervised KPE methods do not rely on training corpora but often exhibit subpar results. Despite the current evidence indicating that KPE based on abstracts surpasses full-text approaches, there is potential for obtaining superior quality keyphrases by integrating algorithms that harness the section structural information within the academic papers.

### 2.2 Section structure classification of academic articles

Previous studies mainly employed manual annotation for section structure classification of academic articles. For instance, De Waard & Kircz (2008) employed a hierarchical structure-function model to manually annotate 13 scientific articles in the biological field. However, this method is time-consuming and labor-intensive, suitable only for processing small datasets. Consequently, many researchers have explored automatic methods for section structure classification. As summarized in Table 3, existing studies on section recognition primarily employ three approaches: rule-based methods, machine learning-based methods and deep-learning methods.

**Table 3.** Summary of representative studies of section structure classification from academic articles

| Categories of methods | Authors | Methods | Main findings |
|---|---|---|---|
| Rule-based methods | *Veyseh et al., 2020* | *ON-LSTM* | The proposed model undergoes extensive analysis and achieves state-of-the-art performance across four benchmark datasets. |
| | *Lopukhina et al., 2021* | *Top-down heuristics methods* | Top-down heuristic processing increases with age, with older adults relying more on semantic heuristics and adolescents not relying on structural heuristics. |
| | *Ma et al., 2022* | *Attention-based BiLSTM* | This study enhances KPE by clarifying the semantic structure of academic literature, demonstrating the effectiveness of chapter titles and relative position as features. |
| ML-based methods | *Jayaram et al., 2020)* | *Naïve Bayes, Decision tree, Maximum entropy* | This work successfully demonstrates that incorporating section structure information from academic articles significantly enhances keyphrase extraction performance. |
| | *Li et al., 2020* | *CRF* | Li et al. proposed an iterative training framework to improve weakly supervised semantic segmentation, achieving state-of-the-art performance on benchmark datasets. |
| | *Zhou et al., 2020* | *Hierarchy-aware global model (HiAGM)* | The hierarchy-aware global model with two variants enhancs hierarchical text classification and advances keyword extraction through improved label dependency modeling and text-label interaction. |
| DL-based methods | *Yu et al., 2017* | *Hybrid Neural Network Hidden Markovl (NN-HMM)* | NN-HMM outperforms traditional HMM in story segmentation, achieving state-of-the-art performance by using deep neural networks to map word distributions into topic posterior probabilities. |
| | *Zhang et al., 2020* | *Graph Neural Networks (GNN)* | This study used GNN for inductive text classification, effectively capturing contextual word relationships within documents, outperforming existing methods on benchmark datasets. |
| | *Wu et al., 2021* | *Transformer* | Hi-Transformer improves keyphrase extraction for long documents by reducing complexity and capturing global context through hierarchical modeling of sentences and documents. |

The rule-based method relies on researchers' analysis and requires the classification of features strongly associated with the corresponding section. These features may include syntactic rules (Veyseh et al., 2020), heuristic phrases (Lopukhina et al., 2021), and contextual information (Ma et al., 2022). For instance, Gupta & Manning (2011) extracted detailed structure information from the titles and abstracts by employing pattern matching, syntactic analysis, and constructing seed rule words. Similarly, Ding et al. (2013) utilized the morphological root form of each word in section names, mapping them to their corresponding sections and enhancing the matching results by adding location-based features. Their approach successfully identified sections such as from academic articles.

The machine learning-based method can be classified into text classification and sequence annotation methods. The text classification method primarily involves the categorization of various text fragments, including sentences, paragraphs, and sections, based on their content into appropriate section structure categories. Sequence labeling methods prioritize text continuity and contextual cues, ideal for annotating diverse sections within articles. Key algorithms include, Naïve Bayes (Jayaram et al., 2020), Conditional Random Fields (CRF, Li et al., 2020). These methods focused on leveraging surface-level features such as text length, keyword frequency, and basic positional information. However, with the advent of deep learning, there has been a notable shift towards more sophisticated models capable of capturing intricate patterns in text data. Recent studies have increasingly employed deep learning models, such as Hidden Markov Models (HMM, Yu et al., 2017)Convolutional Neural Networks (CNNs) and RNNs, to automatically identify the structural elements of academic articles, such as introductions, methods, results, and discussions.

The preceding studies provide methodological directives for the automated classification of section structure within paper. To fully leverage the full-text information and generate higher-quality keyphrases, this paper employs an automatic recognition method to acquire the section structure information of academic articles for the KPE task. Presently, the majority of research efforts are dedicated to enhancing Key Phrase Extraction (KPE) performance through model optimization techniques (Nguyen & Luong, 2010; Zhang et al.). However, it is important to recognize that the quality of the extracted text itself plays a crucial role in determining KPE effectiveness. With this in mind, the current study focuses on extracting distinct keyword sets based on section structure information and utilizes an integration algorithm to generate the final keywords. Previous studies primarily focused on optimizing model selection (Nguyen & Luong, 2010; Zhang et al., 2020) overlooking enhancements in the keyphrase selection mechanism. Taking this as an opportunity, this paper extracts distinct sets of keyphrases based on section structure information and employs an integration algorithm to generate the final keyphrases.

**3. Dataset**

According to the results of the data investigation, commonly used open-source academic paper keyword extraction datasets with full-text content are summarized in Table 4.

**Table 4.** Common Open-Source Full-Text Academic Paper Keyphrase Extraction Datasets

| Year | Dataset | Language | Domain | Samples |
|---|---|---|---|---|
| 2007 | Nguyen2007 | English | Computer Science | 209 |
| 2008 | PubMed | English | Biomedical Sciences | 1320 |
| 2009 | ACM | English | Computer Science | 2304 |
| 2009 | Citeulike-180 | English | Bioinformatics | 183 |
| 2010 | SemEval-2010 | English | Computer Science | 244 |
| 2015 | cacic | Spanish | Computer Science | 888 |
| 2015 | wicc | Spanish | Computer Science | 1640 |

However, since these open-source datasets do not differentiate between sections or paragraphs during text conversion, they pose challenges for subsequent section structure recognition. Therefore, it is necessary to either construct new datasets or obtain structured texts, such as HTML or XML, from existing open-source datasets. We selected papers from three domains for our research: Biomedical science, Computer science, and Library and Information Science (LIS). Generally, there are certain differences in the structural organization of academic papers between natural sciences and social sciences. To ensure the representativeness of the corpus, we chose two disciplines leaning towards natural sciences (Biomedical science and Computer science) and one discipline leaning towards social sciences (LIS).

### 3.1 Construction for Raw Corpora

The primary objective of this study is to explore the impact of the section structure of input texts on KPE performance. Preliminary investigations reveal that journal articles in the three selected domains largely conform to the standard IMRD structure. Based on the research objective, we prioritize constructing a moderately sized raw corpus with well-structured articles. After data collection, the raw data were filtered to exclude academic papers without author keywords. The data were then preprocessed to retain HTML tags for article paragraphs and titles while removing other HTML tags. This process resulted in structured paper datasets for KPE, namely PMC, LIS, and IEEE, with dataset sizes of 1316, 2000, and 2000 papers, respectively.

**(1) PMC Dataset**

This dataset is from Schutz (2008) containing 1323 academic papers with full text, which are available on the PubMed website. After filtering 7 papers without full-text content, the dataset contains 1316 academic papers. PMC primarily comprises papers in the fields of biomedical science[1].

**(2) LIS Dataset**

This is a self-built dataset, containing 2000 papers from library and information science area with full text. These papers are randomly selected from 5 journals: *Aslib Journal of Information Management, Journal of Documentation, Library Hi Tech, Online Information Review, and The Electronic Library*.

**(3) IEEE Dataset**

The papers in the field of computer science are selected from five journals during the period 2018-2020. Specific journal sources are: IEEE Transactions on Pattern Analysis and Machine Intelligence, IEEE Transactions on Cybernetics, IEEE Transactions on Neural Networks and Learning Systems, IEEE Transactions on Evolutionary Computation, and IEEE Transactions on Fuzzy Systems

### 3.2 Construction of KPE Corpora

Motivated by the objectives of this study, it is crucial to first ensure that the section structures of individual input texts are sufficiently uniform. Such consistency is of paramount importance, not only for effectively comparing the performance of KPE in relation to diverse section structure information but also for maintaining the quality of section annotations. Consequently, to facilitate this analysis, we undertook the segmentation and reconstruction of both the training set and the dataset. Initially, based on preliminary investigations into the target domain, we established a section identification framework, which serves as the standard for subsequent section annotations (see Section 4.1).

Then we construct three corpora: *Corpus-PH, Corpus-MH, and Corpus-ML*, representing manually annotated articles with high SSC, model-annotated articles with high SSC, and model-annotated articles with low SSC, respectively. Theoretically, the classification quality of section structure decreases in turn from *Corpus-PH* to *Corpus-MH* and *Corpus-ML*. *Corpus-PH* comprises randomly selected 500 articles from each high SSC corpus, while *Corpus-MH* and *Corpus-ML* consist of the remaining high SSC and low SSC articles, respectively. Due to the substantial volume of full-text academic papers, we decided to randomly shuffle all the data and split it into training, validation, and test sets in an 8:1:1 ratio, instead of adopting ten-fold cross-validation. The number of samples and author's keyphrases for each set are presented in Table 5.

**Table 5.** Number of samples and author's keyphrases of training and test sets in different corpora.

| Corpus | Item | *PMC* | | | *LIS* | | | *IEEE* | | | Total | | |
|---|---|---|---|---|---|---|---|---|---|---|---|---|---|
| | | Train | Valid | Test | Train | Valid | Test | Train | Valid | Test | Train | Valid | Test |
| Corpus-PH | samples | 400 | 50 | 50 | 400 | 50 | 50 | 400 | 50 | 50 | 1,200 | 150 | 150 |
| | keyphrases | 2,064 | 306 | 279 | 2,109 | 219 | 251 | 1,842 | 231 | 212 | 6,015 | 756 | 742 |
| Corpus-MH | samples | 240 | 30 | 30 | 400 | 50 | 50 | 560 | 70 | 70 | 1,200 | 150 | 150 |
| | keyphrases | 1,246 | 228 | 119 | 2,034 | 304 | 289 | 2,634 | 248 | 330 | 5,914 | 780 | 738 |
| Corpus-ML | samples | 412 | 52 | 52 | 800 | 100 | 100 | 640 | 80 | 80 | 1,852 | 232 | 232 |
| | keyphrases | 2,213 | 385 | 264 | 4,022 | 460 | 468 | 2,995 | 380 | 400 | 9,230 | 1225 | 1,132 |

## 4. Methodology

To investigate the potential benefits of utilizing section structure information in KPE, this paper leveraged the structural features and section texts derived from academic articles to enhance KPE performance. The research framework diagram illustrating this approach is depicted in Fig. 1.

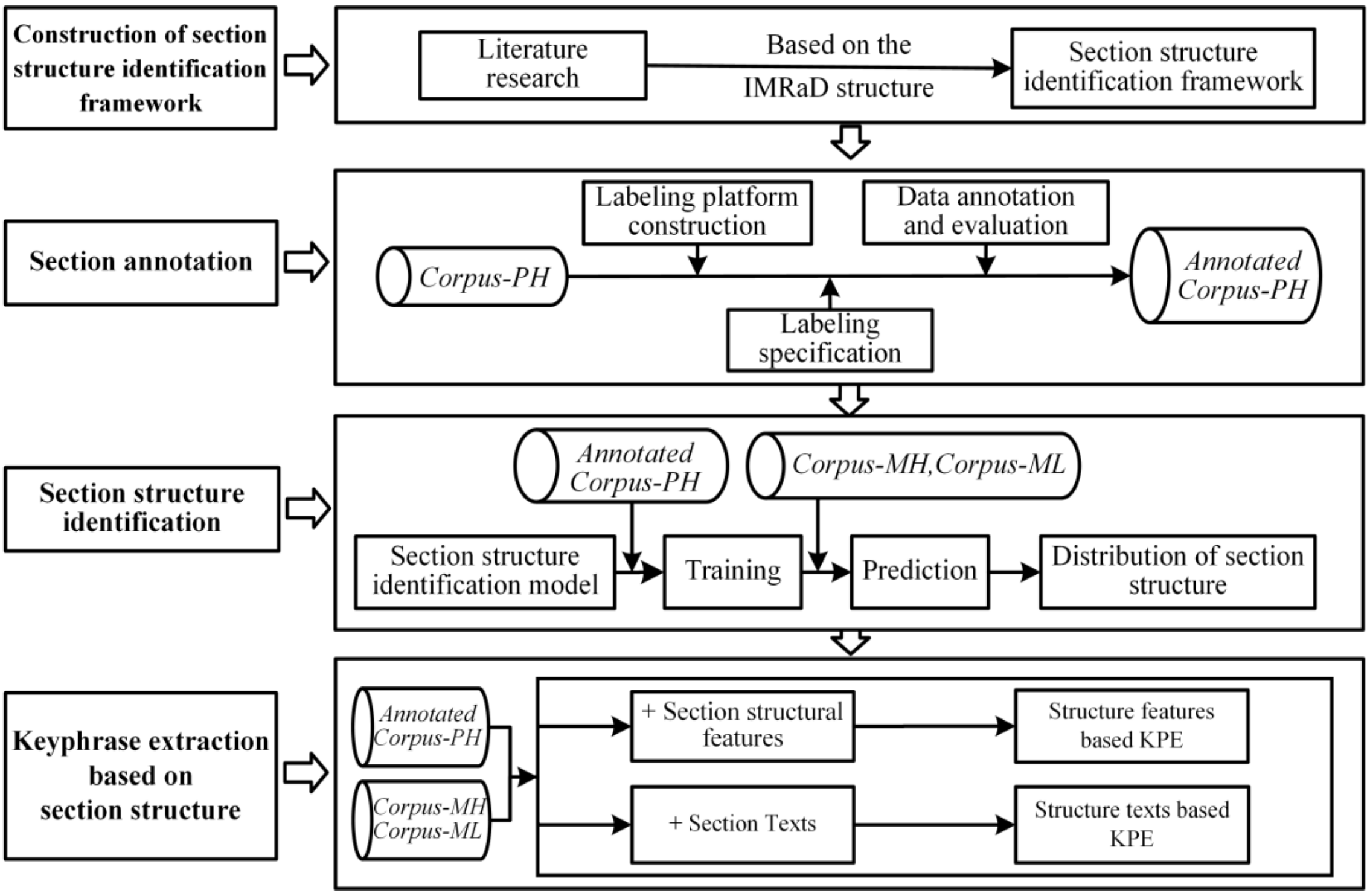

**Fig. 1.** Framework of this study

The framework of this study comprised following four steps. Firstly, a section structure classification framework was established by referencing the IMRaD (Introduction, Methods, Results, and Discussion) (Sollaci & Pereira, 2004) structure of academic papers, which fit the corpus of our study well. In order to deeply examine the impact of the quality of section structure recognition on KPE's performance, we divided all the corpus into three sets of corpora: *Corpus-PH, Corpus-MH, and Corpus-ML*. Detailed introductions are provided in Section 3.2. Subsequently, we manually annotated the section structure of the *Corpus-PH* and employed the Kappa coefficient to evaluate the annotated results. The final step is keyphrase extraction based on section structure. In order to comprehensively assess the influence of section structure information on the KPE task, we incorporate it into the KPE process in the form of structural features and section texts, and evaluate the performance of task respectively.

**4.1. Section annotation and classification**

The study's corpora encompass biomedical science, library and information science, and computer science. Experimental articles in biomedical science and computer science typically feature specialized sections for methods and results. Library and information science, as a social science, often exhibits varying structures but generally includes introduction and conclusion sections. Some articles also have related work sections. Following the IMRaD structure and the section classification system by Ma et al. (2022), this paper classifies sections into six categories: introduction, related work, method, evaluation & result, discussion & conclusion, and others. Common feature words or phrases contained in section headings are extracted as shown in Table 6, and a rule matching method is utilized to annotate the entire corpus.

**Table 6.** List of matching word for section classification

| Category | Matching words |
|---|---|
| Introduction | introduction, motivation, background |
| Related work | overview, literature review, review of literature, related work, prior work |
| Method | method, methodology, theory, model, algorithm, framework, approach, data, dataset, material |
| Evaluation & result | evaluation, result, analysis, experimental result, finding |
| Discussion & conclusion | discussion, conclusion, summary, future work |
| Others | - |

We enlisted four postgraduates specializing in library and information science to manually annotate the section structures of articles in *Corpus-PH*. The annotation yielded an average Kappa coefficient (Carletta, 1996) of 0.938, indicating a high level of consistency among the annotators. While manual annotation method yields high-quality annotations, it is time-consuming and inefficient when dealing with a large number of academic articles. To address this challenge, we propose utilizing a machine learning approach to automatically identify the section structure of academic articles.

### 4.2. Keyphrase extraction based on section structure information

This section provides a comprehensive overview of the experimental configuration employed for keyphrase extraction utilizing section structure. Specifically, we integrate section structure information through structural features and section texts, with the goal of further exploring the influence of section structure on the KPE task.

#### 4.2.1 Construction of keyphrase extraction model

This paper regards the section structure classification of academic articles as a multi-label classification task and designed a deep learning model that integrates the section categories, headings, and non-semantic text features (as shown in Table 7).

The structure of the model is shown in Fig. 2. Firstly, we used CNN (Kim, 2014) to extract the features of the section name and headings of the article, obtaining the corresponding semantic representation vectors $h_1$ and $h_2$. Then, we utilized a CNN or BiLSTM (Graves & Schmidhuber, 2005) or BERT (bidirectional Encoder Representation from Transformers) (Devlin et al., 2019) extracted the features of the section text to obtain the semantic representation vector $h_3$. The semantic representation vectors $h_1$, $h_2$, $h_3$, along with the non-semantic features of the section $T\{T_1, T_2, \ldots, T_{11}\}$, are then concatenated to form the feature representation vector $h_o$.

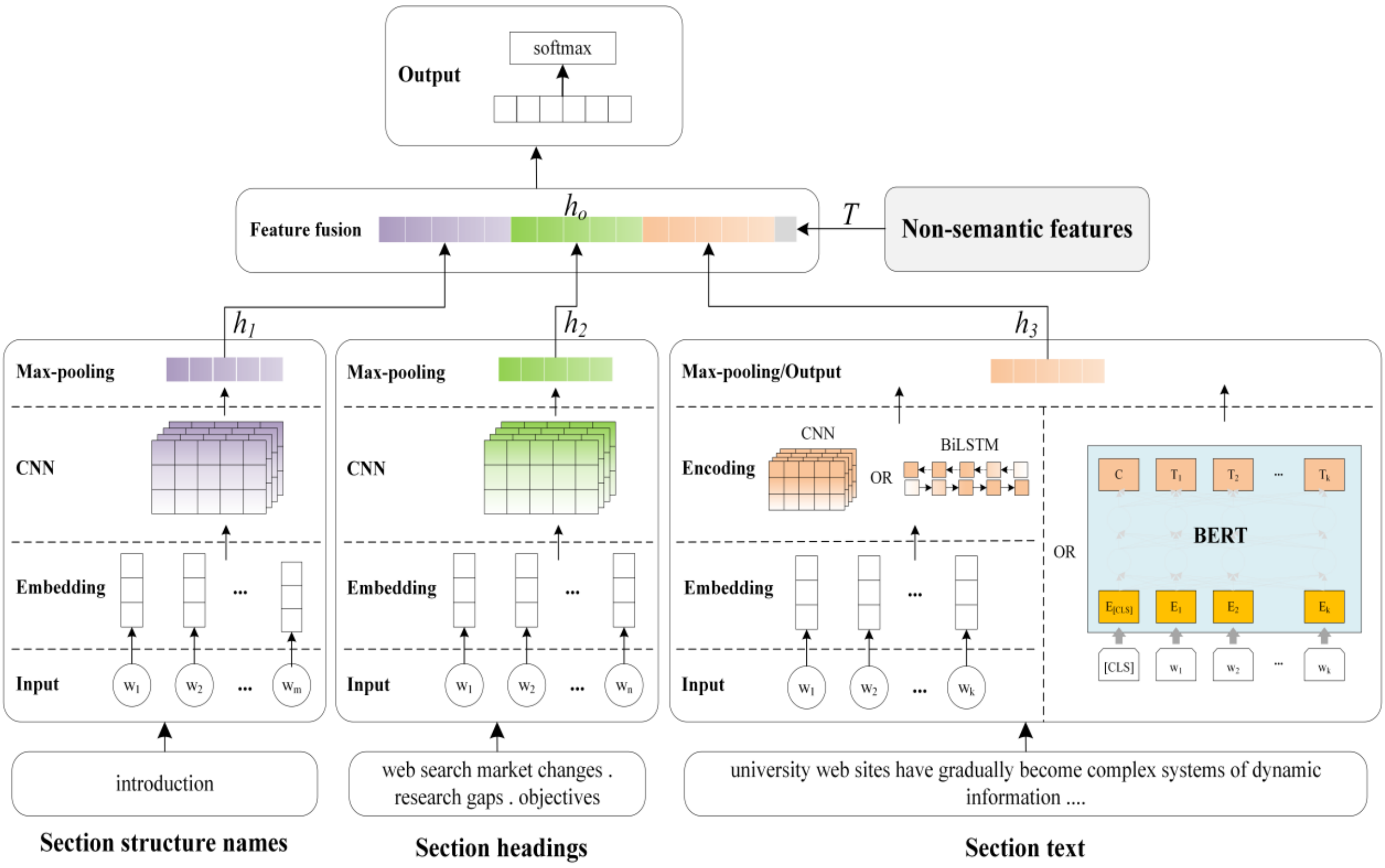

**Fig. 2.** Model structure fusing section name, headings, text, and non-semantic features.

**Table 7.** Non-semantic features used for section structure classification.

| Type | Name | Description | Dimension |
|---|---|---|---|
| Dataset properties | $T_1$ | Dataset ID of the article containing the section. | 1 |
| | $T_2$ | Journal ID of the article containing the section. | 1 |
| Mention statistics | $T_3$ | Number of citations mentioned in section text. | 1 |
| | $T_4$ | Number of footnotes mentioned in section text. | 1 |
| | $T_5$ | Number of figures mentioned in section text. | 1 |
| | $T_6$ | Number of tables mentioned in section text. | 1 |
| | $T_7$ | Number of formulas mentioned in section text. | 1 |
| Text length | $T_8$ | Number of paragraphs in section text. | 1 |
| | $T_9$ | Number of sentences in section text. | 1 |
| | $T_{10}$ | Number of words in section text. | 1 |
| Section locations | $T_{11}$ | Location of the section in full text. | 1 |

According to different models for semantic coding of section text, we constructed five models based on CNN, BiLSTM, BERT, BiLSTM-Att, and BERT-Att, respectively. The BiLSTM-Att model integrates a BiLSTM network with additive attention to extract features from chapter content, while other structures remain unchanged (same as BERT-Att). Naive Bayes (NB), SVM, and random forest (RF) were used as baseline models, and P, R, and $F_1$ values were used as evaluation metrics. Each sample includes section titles, sub-section titles, content of sections, and non-semantic features.

#### 4.2.2. KPE based on structural features

In academic articles, different sections contribute to keyphrases at different rates (Nguyen & Kan, 2007), leading to diverse distributions of keyphrases across sections. To leverage this distribution in KPE, we constructed ten baseline features and seven structural features to analyze their impact on KPE models. We selected SVM (Zhang et al., 2006) for its ability to handle high-dimensional features and CRF (Zhang et al., 2008) for its strength in addressing sequence dependencies. The details of these features are in Table 8. Notably, the semantic similarity between words and text ($F_8$) is calculated using cosine similarity based on GloVe-trained (Pennington et al., 2014) semantic vectors. 'Dimension' refers to the number or count of features. For instance, the dimension of $F_1$ is 1, indicating that there is only one dimension associated with it. It means that this feature considers information related to only one aspect, which is the position of the word as the first occurrence in the text.

In this paper, the candidate phrases are noun phrases from the titles and abstracts of academic articles. Subsequently, SVM and CRF models combined with baseline features were employed to evaluate the fundamental performance of the models. Finally, different structural features were added to compare the features of section structure on KPE performance.

**Table 8. Base and structural features for KPE.**

| Type | Name | Description | Dimension |
|---|---|---|---|
| Baseline features | $F_1$ | First position of the word in the text | 1 |
| | $F_2$ | Last position of the word in the text | 1 |
| | $F_3$ | Span of the word in the text, equal to $\boldsymbol{F_2 - F_1}$ | 1 |
| | $F_4$ | Word frequency of the word in the text | 1 |
| | $F_5$ | Part-of-Speech of the word in the text | 1 |
| | $F_6$ | Number of words composing the phrase | 1 |
| | $F_7$ | Number of characters composing the word | 1 |
| | $F_8$ | Semantic similarity between the word and text | 1 |
| | $F_9$ | TF*IDF value of the word in the text | 1 |
| | $F_{10}$ | TextRank value of the word in the text | 1 |
| Structural features | $F_{11}$ | First position of the word in different section texts | 5 |
| | $F_{12}$ | Last position of the word in different section texts | 5 |
| | $F_{13}$ | Span of the word in different section texts, equal to $\boldsymbol{F_{12} - F_{11}}$ | 5 |

| | | |
|---|---|---|
| $F_{14}$ | Word frequency of the word in different section texts | 5 |
| $F_{15}$ | Semantic similarities between the word and different section texts | 5 |
| $F_{16}$ | TF*IDF values of the word in different section texts | 5 |
| $F_{17}$ | TextRank values of the word in different section texts | 5 |

**Note:** Dimension refers to the number or count of features. For instance, the dimension of F1-score is 1, indicating that there is only one dimension associated with it.

#### 4.2.3. KPE based on section texts

**Table 9. Description of the keyphrase integration algorithm.**

| | |
|---|---|
| **Keyphrase Integration Algorithm**: Integrating the extraction results from each section | |
| **Input:** For an article, keyphrases are extracted from "title+introduction", "title+related work", "title+method", "title+evaluation & result", and "title+discussion & conclusion" to get different keyphrase sets $KS_i$. | |
| **Ouput:** Keyphrase integration result. | |
| Step1 | In each keyphrase set $KS_i$, $N$ words in the top rank are selected by the importance or probability value of the keyphrase to obtain the keyphrase set $RS_i$. |
| Step2 | Merge all keyphrase sets $RS_i$ to get the set $AS$. |
| Step3 | In the set $AS$, each keyphrase is sorted by word frequency and obtain the keyphrase set $SS$. If the word frequency of keyphrases is the same, they would be sorted by importance or probability value. |
| Step4 | In the keyphrase set $SS$, select the top keyphrases as the keyphrase integration result. |

To explore the effect of section text on KPE, this paper employed various section texts as input corpora for the model. The models employed include TF*IDF (Salton & Buckley, 1988), TextRank (Mihalcea & Tarau, 2004), SVM, CRF, BiLSTM-CRF (Al-Zaidy et al., 2019), and BERT-BiLSTM-CRF (Hien et al., 2021). Subsequently, the keyphrase integration algorithm is utilized to consolidate the results from each section into the final keyphrases. The steps of the keyphrase integration algorithm are presented in Table 9. The importance or probability value of a keyphrase refers to the TF*IDF and TextRank values in their corresponding models. For the SVM model, it corresponds to the probability value of the phrases. As for the CRF, BiLSTM-CRF, and BERT-BiLSTM-CRF models, it represents the average probability of tags indicating keyphrases. For instance, let's consider the keyphrase "pedestrian lane detection" with the probability of each word $\{\mathrm{B} = 0.8,\ \mathrm{I} = 0.6,\ \mathrm{I} = 0.4\}$. The value of the keyphrase would be calculated as $(0.8 + 0.6 + 0.4)/3 = 0.6$.

## 5. Result analysis

This section analyzes experimental results to evaluate KPE performance and the impact of structural information. Section 5.1 introduces the metrics for evaluating the performance of KPE. Section 5.2 examines the effects of structural features and section texts on KPE. Section 5.3 quantifies the influence of section structure recognition quality using performance differences (CS-AB, CS-FT). Section 5.4 presents two case studies validating the impact of section structure and the proposed integration algorithm using the BERT-BiLSTM+CRF model.

### 5.1 Evaluation metrics for KPE

This paper evaluated the KPE performance by calculating the $F_1@K$ value based on exact string matching. Before conducting the exact string matching, the stems of the predicted keyphrase and the author-provided keyphrase are extracted. The $F_1@K$ value is calculated as follows:

$$F_1@K = 2 \times \frac{P@K \times R@K}{P@K + R@K} \times 100\% \tag{1}$$

$$P@K = \frac{Y_{pred_k} \cap Y_{gold}}{K} \times 100\% \tag{2}$$

$$R@K = \frac{Y_{pred_k} \cap Y_{gold}}{|Y_{gold}|} \times 100\% \tag{3}$$

Where, $Y_{pred}$ represents the keyphrase sequence predicted by the model, while $Y_{gold}$ denotes the author-provided keyphrase sequence. $Y_{pred_k}$ indicates the top $K$ phrases in the predicted sequence (Refer to Section 4.4 for the details on probability value calculations). K is set to 3, 5, and 10. The Wilcoxon signed-rank test (Woolson, 2008) assesses the statistical significance of extraction results based on F_1@K values, with a significance level of 0.05.

### 5.2 Impact of structural information on KPE models

Table 10 compares the performance of section structure classification models, showing that the CNN-based model excels with an F1-score of 92.21%, followed by the BERT-Att model at 91.22%. The NB model performs worst, scoring 76.62%. Deep learning models overall outperform traditional machine learning models. Most models perform better on PMC and IEEE datasets compared to LIS. This disparity is likely due to the unclear text structures in LIS articles, which complicate accurate section structure identification. The CNN-based model captures local patterns and dependencies within text more accurately through convolution operations. Compared to BiLSTM, CNNs excel at extracting local features and are advantageous in classifying sections, where much information is determined by local context. Thus, the CNN-based model is recommended for predicting section structures in articles.

**Table 10**. Performance of various section structure classification models.

| Models | *PMC* | | | *LIS* | | | *IEEE* | | | *Overall* | | |
|---|---|---|---|---|---|---|---|---|---|---|---|---|
| | **P** | **R** | **F1** | **P** | **R** | **F1** | **P** | **R** | **F1** | **P** | **R** | **F1** |
| Naïve Bayes (NB) | 76.10 | 85.88 | 77.66 | 74.68 | 74.71 | 72.78 | 78.89 | 81.63 | 77.13 | 77.28 | 86.48 | 76.62 |
| SVM | 82.41 | 85.15 | 81.81 | 81.27 | 78.13 | 78.69 | 83.28 | 80.05 | 79.79 | 89.68 | 80.04 | 82.42 |
| Random Forest (RF) | 84.36 | 84.56 | 83.90 | **84.79** | **84.61** | **84.66** | 90.87 | 89.48 | 90.06 | 93.75 | 87.54 | 89.39 |
| CNN | 88.46 | 91.09 | 88.33 | 84.09 | 83.82 | 83.87 | 93.20 | 93.01 | 93.08 | 96.41 | **90.15** | **92.21** |
| BiLSTM | 86.70 | 87.73 | 86.04 | 83.98 | 83.69 | 83.77 | **93.74** | **93.34** | **93.52** | 93.27 | 88.49 | 90.09 |
| BERT | 80.18 | 85.94 | 81.75 | 84.42 | 83.68 | 83.90 | 90.29 | 89.88 | 90.06 | 89.99 | 86.12 | 87.36 |
| BiLSTM-At | 84.97 | 87.62 | 84.84 | 84.15 | 84.17 | 84.11 | 89.91 | 89.85 | 89.85 | 93.01 | 87.51 | 89.03 |
| BERT-Att | **90.10** | **94.32** | **90.48** | 84.32 | 83.88 | 83.99 | 90.29 | 89.59 | 89.89 | **96.65** | 88.87 | 91.22 |

**Note**: The bold font represents the model that achieves the best performance on a metric.

The KPE results based on section structure information are divided into two parts: analyzing the effect of structural features on KPE model performance and examining the impact of different section texts on KPE performance and the integrated results.

### 5.2.1 Impact of structural features on KPE models

Building upon the baseline features, this paper employed SVM and CRF as KPE models respectively. The source text is the "title+abstract" of articles. We subsequently added different structural features to investigate their enhancement. The results are presented in Fig. 3. where "+$F_{11}$" represents the addition of the structural feature F11 to the model, with analogous meanings for other symbols.

The results in Fig. 3 demonstrate that some of the structural features have an enhancing effect on the performance of the KPE model. Structural features $F_{13}$ and $F_{17}$ significantly enhance KPE model performance, while $F_{16}$ performs the worst and has almost no positive effect. $F_{13}$ exerts a greater influence on SVM, while $F_{17}$ manifests a more pronounced effect on CRF. $F_{11}$ and $F_{12}$ respectively denote partial features of word distribution across different sections, thus their impact on the model enhancement is moderate. In contrast, $F_{13}$ can precisely reflect the span of words within the text, hence yielding a more favorable effect on improvement of model performance.

However, the same structural feature may have different effects on SVM and CRF. For instance, the effect of $F_{14}$ on SVM is more pronounced, while $F_{15}$ has a more significant effect on CRF. Moreover, the same feature can also have varied performance in different corpora. The structural feature $F_{12}$ enhances CRF's performance in *Corpus-MH* and *Corpus-ML*, but reduces CRF's performance in *Corpus-PH*. Overall, different structural features can facilitate KPE, but their performance varies. Structural features are valuable components of section information in KPE research.

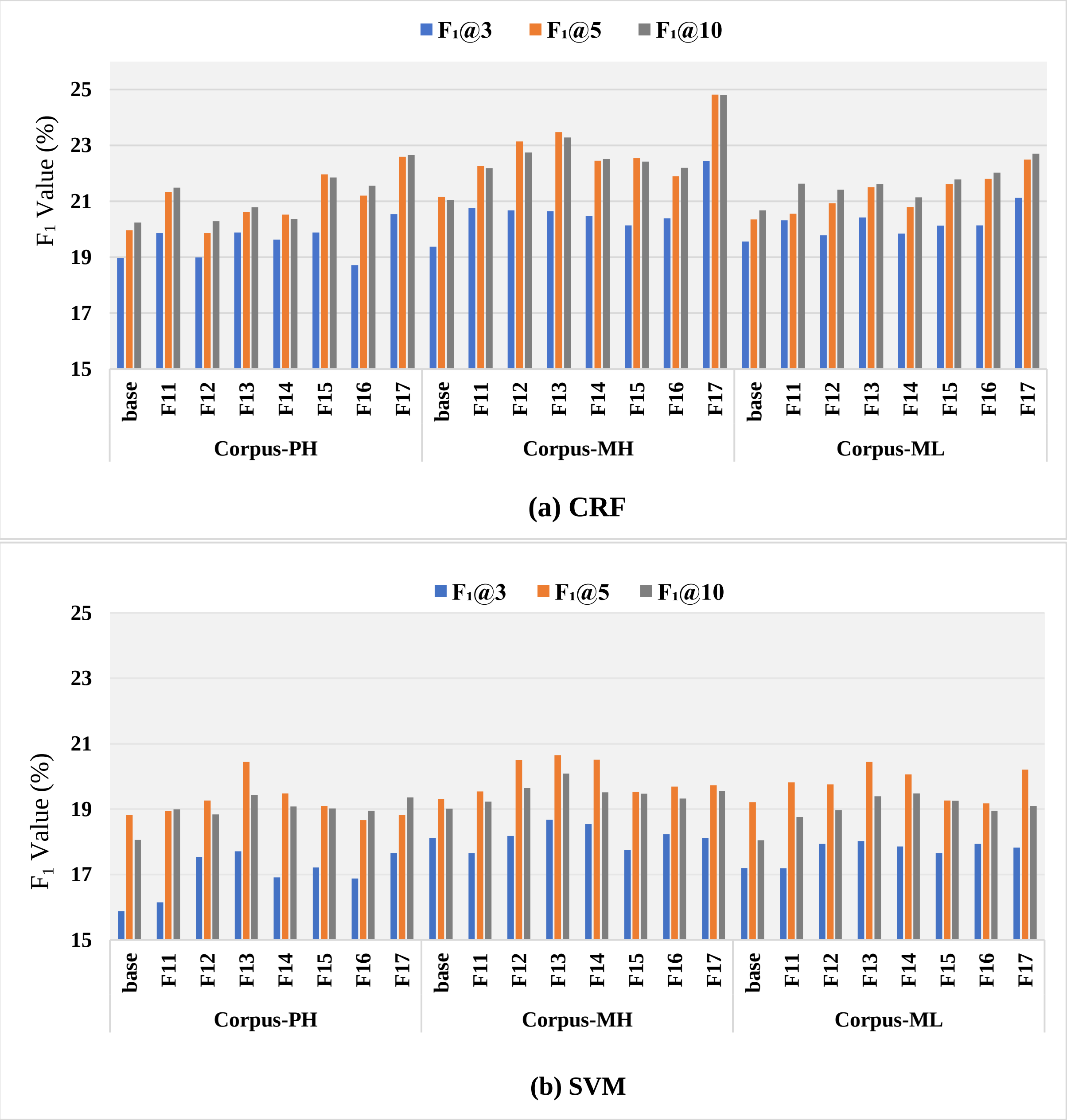

**Fig. 3.** Performance changes of SVM and CRF on different structural features

### 5.2.2 Impact of section text on KPE

To investigate the effectiveness of this method, the full text was divided into five parts according to the classification framework proposed above. Keyphrases were extracted from each part, and the extraction results WS, SS and CS were obtained through integrating the keyphrases according to the number of words, sentences and sections in each article respectively. The abbreviations used are defined in Table 11. As depicted in Fig. 4, Fig. 5, and Fig. 6, the performance of *FT* as the input corpus for each KPE model is generally poor. Similarly, relying on *IN, RW, MD, ER*, and *DC* does not yield satisfactory results. However, the application of the keyphrase integration algorithm notably improves performance. Specifically, when TextRank and SVM models are employed to extract keyphrases in *Corpus-PH*, the keyphrase integration result *CS* outperforms *AB* and *FT* across various domain datasets and evaluation metrics. Although it is evident that both WS and SS also demonstrate superior performance, *CS* exhibits the best overall performance.

**Table 11.** The meaning of section texts abbreviation and the integration results abbreviation

| **Type** | **Abbreviation** | **Meaning** |
|---|---|---|
| *Section texts* | *AB* | Title and Abstract |
| | *FT* | Title and full text the article |
| | *IN* | Title and Introduction |
| | *RW* | Title and Introduction of the article. |
| | *MD* | Title and method of the article. |
| | *ER* | Title and Evaluation & result |
| | *DC* | Title and Discussion & conclusion |
| *Integration result* | *WS* | according to the number of words |
| | *SS* | according to the number of sentences |
| | *CS* | according to the number of sections |

When comparing the performance of various KPE models, the TF*IDF and TextRank models exhibit poor performance. The BERT-BiLSTM-CRF model yields the best performance, while the performance of the BiLSTM-CRF model is significantly reduced, even falling below that of the SVM and CRF models. This is attributed to BERT's robust capability of acquiring dynamic word vectors, surpassing the static word embedding layer in the BiLSTM-CRF model. With limited training data, the BERT-BiLSTM-CRF model excels, while the BiLSTM-CRF model underperforms due to insufficient learning.

**(a) TF*IDF**

| | | AB | FT | IN | RW | MD | ER | DC | WS | SS | CS |
|---|---|---|---|---|---|---|---|---|---|---|---|
| PMC-1316 | F1@3 | 10.82 | 12.19 | 12.72*† | 13.9*† | 8.86 | 9.03 | 13.84† | 13.79† | 12.68*† | 14.3*† |
| | F1@5 | 12.51 | 11.81 | 14.66*† | 15.2*† | 9.79 | 9.76 | 11.38 | 16.08*† | 15.27*† | 15.77* |
| | F1@10 | 14.06 | 10.18 | 13.83 | 15.04* | 9.93 | 8.78 | 10.94† | 17.82 | 17.74 | 17.62* |
| LIS-2000 | F1@3 | 14.65 | 10.78 | 14.94 | 14.29 | 7.87 | 8.54 | 10.53 | 12.86† | 12.86† | 15.29† |
| | F1@5 | 17.53 | 10.7 | 15.26 | 14.47 | 11.77† | 9.21 | 13.06† | 15.98 | 15.26 | 18.85* |
| | F1@10 | 16.32 | 12.83 | 14.14† | 12.7 | 11.53 | 10.4 | 11.41 | 18.49* | 19.15 | 19.23 |
| IEEE-2000 | F1@3 | 8 | 3.67 | 6.1 | 6.72 | 4.42† | 2.74† | 6.62 | 8 | 6.33 | 8.87 |
| | F1@5 | 9.15 | 5.39 | 8.16 | 9.29† | 3.93† | 3.55 | 7.32† | 10.25 | 8.92† | 10.12 |
| | F1@10 | 10.03 | 5.14 | 9.15 | 9.27 | 5.03 | 4.56 | 7.46† | 11.67* | 10.54* | 12.5* |

**(b) TextRank**

| | | AB | FT | IN | RW | MD | ER | DC | WS | SS | CS |
|---|---|---|---|---|---|---|---|---|---|---|---|
| PMC-1316 | F1@3 | 12.72 | 10.34 | 11.77† | 7.56 | 3.85 | 8.29 | 12.06† | 12.89*† | 12.82*† | 12.94 |
| | F1@5 | 14.19 | 11.51 | 14.74* | 8.63 | 6.79 | 7.87 | 12.5† | 14.78*† | 13.81† | 12.69† |
| | F1@10 | 13.92 | 9.84 | 12.57 | 12.49† | 6.67 | 7.56 | 10.72† | 15.94* | 14.3*† | 13.38 |
| LIS-2000 | F1@3 | 13.26 | 11.15 | 12.55† | 13.35*† | 7.47 | 9.21 | 11.73† | 14.59*† | 13.92*† | 16.18*† |
| | F1@5 | 14.31 | 12.39 | 13.63† | 14.87*† | 9.96 | 10.76 | 12.59† | 18.37 | 18.37 | 18.75* |
| | F1@10 | 14.2 | 12.05 | 14.56*† | 12.39† | 10.54 | 10.79 | 11.61 | 17.9 | 17.1 | 19.25 |
| IEEE-2000 | F1@3 | 7.94 | 5.28 | 6.86† | 4.05 | 4.29 | 3.1 | 6.42† | 6.46† | 6.76† | 9.99* |
| | F1@5 | 9.62 | 6.79 | 8.48† | 7.39† | 3.88 | 3.75 | 7.11† | 7.55† | 10.19*† | 10.42* |
| | F1@10 | 9.1 | 6.37 | 8.01† | 9.78* | 3.87 | 4.33 | 6.27 | 10.37* | 10.62* | 11.34* |

**(c) SVM**

| | | AB | FT | IN | RW | MD | ER | DC | WS | SS | CS |
|---|---|---|---|---|---|---|---|---|---|---|---|
| PMC-1316 | F1@3 | 11.01 | 11.88 | 10.96† | 9.64 | 11.58 | 9.33† | 16.83*† | 16.76*† | 15.95*† | 17.46 |
| | F1@5 | 15.29 | 13.82 | 13.43 | 11.99 | 12.1 | 11.45 | 16.35*† | 17.09*† | 17.93*† | 18.48† |
| | F1@10 | 16.03 | 13.97 | 15.06† | 15.5† | 12.73 | 10.83 | 15.55† | 16.95† | 16.28† | 17.02† |
| LIS-2000 | F1@3 | 16.2 | 9.85 | 15.83*† | 15.19 | 12.41 | 8.11 | 11.18† | 13.23† | 11.74† | 13.7*† |
| | F1@5 | 17.1 | 11.77 | 17.4† | 17.22 | 12.58 | 12.19 | 13.56† | 16.57† | 16.37† | 18.02 |
| | F1@10 | 17.56 | 12.72 | 18.22 | 16.1† | 13.75 | 13.33 | 14.24† | 15.2† | 15.17† | 18.12 |
| IEEE-2000 | F1@3 | 14.52 | 10.29 | 14.4 | 14.79*† | 10.71 | 9.94 | 14.29† | 16.07† | 16.52† | 16.44† |
| | F1@5 | 16.99 | 14.45 | 17.93 | 15.13† | 11.33 | 11.02 | 14.99† | 17.39† | 17.09† | 17.89† |
| | F1@10 | 16.07 | 13.91 | 16.72† | 14.91 | 12.96 | 11.9 | 14.06† | 16.05† | 15.78† | 16.47 |

**(d) CRF**

| | | AB | FT | IN | RW | MD | ER | DC | WS | SS | CS |
|---|---|---|---|---|---|---|---|---|---|---|---|
| PMC-1316 | F1@3 | 14.14 | 9.78 | 16.5* | 7.32 | 11.39 | 12.02 | 11.79† | 18.51* | 17.53* | 19.13 |
| | F1@5 | 16.48 | 12.52 | 18.53* | 7.23 | 12.1 | 16.53 | 16.93*† | 18.79* | 18.1* | 20.01*† |
| | F1@10 | 16.78 | 16.19 | 18.48*† | 7.23 | 13.95 | 14.09 | 17.63*† | 16.72† | 17.57*† | 19.36*† |
| LIS-2000 | F1@3 | 19.28 | 17.39 | 20.66* | 14.87 | 18.58† | 18.2† | 16.77 | 21.62* | 22.97 | 21.98* |
| | F1@5 | 19.6 | 22.04 | 20.58*† | 16.37 | 19.77† | 17.24 | 18.92 | 23.82† | 23.76 | 22.66† |
| | F1@10 | 19.87 | 21.81 | 20.18*† | 16.55 | 19.62 | 16.87 | 18.49 | 20.64*† | 21.76*† | 21.77* |
| IEEE-2000 | F1@3 | 21.67 | 14.01 | 21.01* | 17.13† | 12.39 | 12.17 | 16.65† | 18.81† | 17.2† | 23.56* |
| | F1@5 | 22.98 | 13.79 | 21.73 | 16.11† | 13.57 | 11.27 | 17.5† | 20.55 | 21.89 | 24.05* |
| | F1@10 | 23.16 | 15.59 | 20.88 | 17.28† | 13.31 | 12.07 | 17.78† | 19.12 | 18.3† | 20.97 |

**(e) BiLSTM+CRF**

| | | AB | FT | IN | RW | MD | ER | DC | WS | SS | CS |
|---|---|---|---|---|---|---|---|---|---|---|---|
| PMC-1316 | F1@3 | 7.08 | 6.24 | 10.71† | 9.48† | 7.49 | 6.02 | 8.29† | 10.1 | 11.64* | 12.76 |
| | F1@5 | 7.98 | 8.21 | 10.28† | 9.44† | 7.04 | 6.09 | 10.48† | 11.98 | 11.98 | 12.85† |
| | F1@10 | 8.12 | 7.83 | 11.59† | 9.69† | 7.22 | 5.97 | 9.89† | 11.22 | 10.86 | 11.94† |
| LIS-2000 | F1@3 | 16.44 | 19.58 | 19.68† | 9.61 | 15.49† | 15.97 | 13.23† | 22.43* | 20.43* | 23.63 |
| | F1@5 | 18.7 | 19.92 | 22.92*† | 11.15 | 15.05† | 14.88† | 13.59† | 20.19* | 21.88* | 23.08* |
| | F1@10 | 18.8 | 17.84 | 23.41*† | 12.17 | 14.31† | 14.56† | 13.38† | 17.96* | 19.31 | 19.28* |
| IEEE-2000 | F1@3 | 15.64 | 15.37 | 14.79† | 13.58† | 11.07 | 11.56 | 11.67 | 14.49 | 14.39† | 19.93* |
| | F1@5 | 14.92 | 16.06 | 14.04† | 12.39† | 10.13 | 9.63 | 11.81† | 15.1 | 15.34* | 17.79 |
| | F1@10 | 15.18 | 14.74 | 16.81*† | 14.86† | 10.88 | 11.25† | 13.81† | 13.46† | 14.19† | 16.19 |

**(f) BERT-BiLSTM-CRF**

| | | AB | FT | IN | RW | MD | ER | DC | WS | SS | CS |
|---|---|---|---|---|---|---|---|---|---|---|---|
| PMC-1316 | F1@3 | 21.56 | 15 | 24.65† | 17.41† | 21.2 | 16.83† | 18.12† | 26.22*† | 24.71* | 26.34 |
| | F1@5 | 23.73 | 15.52 | 23.28 | 17.36† | 19.55 | 19.37† | 20.61† | 27.1*† | 26.1† | 28.13 |
| | F1@10 | 24.55 | 14.57 | 23.78 | 17.4† | 20.69 | 17.08† | 19.2† | 23.34† | 24.21† | 25.14 |
| LIS-2000 | F1@3 | 21.04 | 14.44 | 23.58† | 16.13† | 22.07† | 16.14† | 18.33† | 28.44*† | 21.22*† | 28.94 |
| | F1@5 | 24.32 | 14.81 | 22.49† | 17.72† | 22.45† | 17.74† | 19.35† | 24.67* | 25.1† | 27.55 |
| | F1@10 | 24.4 | 17.36 | 23.2† | 17.12 | 20.65† | 16.71 | 20.05† | 24.43† | 21.76† | 24.15 |
| IEEE-2000 | F1@3 | 23.77 | 17.15 | 20.91† | 16.24† | 17.76 | 14.53 | 20.09† | 22.65† | 22.26† | 26.68 |
| | F1@5 | 23.3 | 18.82 | 21.38† | 17.9† | 16.19 | 15.03 | 21.49† | 26.43† | 26.1† | 26.84 |
| | F1@10 | 26.2 | 16.36 | 19.13† | 17.55† | 12.72 | 14.86 | 20.82† | 21.69† | 21.04† | 22.04 |

**Note:** The symbols "*" and "†" indicate that the KPE result from an input corpus is significantly higher than that from "*AB*" and "*FT*", respectively. Same for the Fig. 5 and Fig. 6. Bold annotations indicate the best scores for each input text, and underlines presents the best score for each KPE model. Data highlighted in red indicate superior performance, white signifies moderate performance, while those in blue indicates inferior performance.

**Fig. 4. Performance of KPE models on *Corpus-PH*.**

**(a) TF*IDF**

| | | AB | FT | IN | RW | MD | ER | DC | WS | SS | CS |
|---|---|---|---|---|---|---|---|---|---|---|---|
| PMC-1316 | F1@3 | 11.73 | 13.5 | 11.82*† | 15.15† | 8.71 | 8.19 | 8.65† | 15.54* | 17.84* | 20.74* |
| | F1@5 | 12.99 | 13.28 | 12.48* | 16.2† | 9.3 | 8.25 | 10.85† | 18.31* | 19.83 | 22.42* |
| | F1@10 | 12.7 | 11.17 | 12.99 | 17.19 | 8.68 | 8.64 | 13.64*† | 19.38 | 21.51 | 20.33 |
| LIS-2000 | F1@3 | 13.18 | 8.65 | 10.84† | 11.83 | 9.94† | 6.12 | 10.61† | 11.17† | 11.12† | 13.64* |
| | F1@5 | 13.79 | 7.59 | 11.55† | 12.28 | 9.69 | 8.09 | 11.8† | 16.06* | 16.1* | 15.36* |
| | F1@10 | 13.65 | 6.01 | 11.65 | 11.13 | 8.39† | 8.36† | 10.75† | 15.66* | 16.6* | 15.99* |
| IEEE-2000 | F1@3 | 6.96 | 3.41 | 6.76† | 6.16 | 4.57 | 3.54 | 6.76† | 7.18† | 6.06† | 8.91 |
| | F1@5 | 8.08 | 5.74 | 8.41† | 7.95 | 5.36 | 4.7 | 8.64† | 8.07† | 8.41† | 10.97* |
| | F1@10 | 10.94 | 5.36 | 8.93 | 9.09 | 5.2 | 5.05 | 9.07 | 9.39 | 9.14 | 11.97 |

**(b) TextRank**

| | | AB | FT | IN | RW | MD | ER | DC | WS | SS | CS |
|---|---|---|---|---|---|---|---|---|---|---|---|
| PMC-1316 | F1@3 | 13.93 | 8.88 | 10.96† | 10.6† | 3.56 | 8.83 | 8.83 | 13.85 | 13.97*† | 18.95 |
| | F1@5 | 13.15 | 8.63 | 11.51† | 14.62*† | 4.25 | 8.91† | 10.85† | 17.39* | 17.46* | 17.71* |
| | F1@10 | 11.49 | 8.91 | 10.47† | 17.34* | 4.66 | 7.92 | 9.57† | 16.62 | 17.33 | 17.17 |
| LIS-2000 | F1@3 | 11 | 11.94 | 14.66*† | 10.57 | 8.28 | 6.87 | 11.58* | 13.54*† | 13.09*† | 16.03 |
| | F1@5 | 12.6 | 13.7 | 17.34* | 11.73 | 7.21 | 9.48 | 12.78* | 15.62*† | 14.89*† | 17.37† |
| | F1@10 | 12.39 | 11.7 | 15.65* | 11.55 | 7.14 | 7.19 | 11.58 | 16.41 | 15.37† | 16.63 |
| IEEE-2000 | F1@3 | 8.72 | 4.04 | 6.69 | 8.32 | 3.18 | 4.83† | 8.49 | 7.44 | 8.13 | 11.42* |
| | F1@5 | 10.44 | 8.78 | 9.45† | 11.88* | 3.39 | 5.17 | 8.9† | 10.98*† | 10.42† | 13.46* |
| | F1@10 | 12.02 | 7.1 | 7.93† | 11.85 | 4.51 | 4.48 | 8.41† | 11.01 | 10.23 | 13.78* |

**(c) SVM**

| | AB | FT | IN | RW | MD | ER | DC | WS | SS | CS | |
|---|---|---|---|---|---|---|---|---|---|---|---|
| PMC-1316 | F1@3 | 15.25 | 12.36 | 16.03*† | 12.64 | 10.58 | 13.5 | 16.83 | 14.76*† | 13.95*† | 17.46 |
| | F1@5 | 16.97 | 15.81 | 16.36† | 13.99† | 10.1 | 12.81 | 18.35† | 15.3* | 15.25† | 18.85† |
| | F1@10 | 19.16 | 14.94 | 16.62 | 17.34† | 9.73 | 11.85 | 16.53 | 15.59 | 16.28† | 19.02† |
| LIS-2000 | F1@3 | 11.73 | 9.38 | 14.68 | 13.9† | 13.18 | 11.94 | 12.94† | 15.45† | 15.85 | 16.47* |
| | F1@5 | 16.51 | 11.05 | 16.45 | 10.46 | 13.78 | 13.58 | 11.56 | 14† | 17.57 | 18.4 |
| | F1@10 | 15.97 | 11.22 | 15.95† | 11.44† | 13.81 | 12.1 | 12.79 | 16.24† | 17.58 | 18.41 |
| IEEE-2000 | F1@3 | 18.48 | 14.55 | 14.1 | 16.6* | 10.45 | 10.25 | 15.72† | 15.87 | 16.17 | 18.76† |
| | F1@5 | 19.73 | 16.5 | 15.37 | 16.09* | 11.66 | 11.04 | 16.66 | 18.12† | 17.48 | 18.54† |
| | F1@10 | 17.43 | 16.74 | 16.34 | 16.19 | 12.61 | 11.34 | 15.44† | 17.72† | 17.54 | 18.3 |

**(d) CRF**

| | | AB | FT | IN | RW | MD | ER | DC | WS | SS | CS |
|---|---|---|---|---|---|---|---|---|---|---|---|
| PMC-1316 | F1@3 | 12.87 | 11.02 | 12.86† | 6.28 | 10.22 | 8.13 | 16.77*† | 19.01* | 18.31* | 19.35 |
| | F1@5 | 12.95 | 13.83 | 15.1*† | 7.06 | 11.13 | 9.18 | 17.8*† | 18.3*† | 18.76† | 19.81*† |
| | F1@10 | 12.83 | 15.21 | 14.49* | 7.06 | 12.69 | 9.04 | 18.49*† | 18.07*† | 19.17*† | 20.47*† |
| LIS-2000 | F1@3 | 21.4 | 17.06 | 22.49* | 14.72 | 15.52 | 14.91 | 18.5† | 21.81* | 22.75* | 22.9* |
| | F1@5 | 22.71 | 19.16 | 22.47* | 17.12 | 18.02 | 16.54 | 19.11 | 23.4*† | 22.44*† | 25.23* |
| | F1@10 | 22.52 | 18.66 | 22.44* | 16.32 | 18.97† | 17.8 | 19.65† | 21.93† | 21.42 | 23.73 |
| IEEE-2000 | F1@3 | 25.69 | 18.25 | 20.6† | 14.54 | 16.73 | 18.08 | 18.68† | 24.13* | 25.45 | 26.01 |
| | F1@5 | 26.03 | 19.57 | 22.32*† | 15.42 | 19.06 | 18.27 | 18.84 | 25.97* | 27.24 | 26.52 |
| | F1@10 | 26.2 | 18.99 | 22.68* | 15.37 | 19.71† | 19.3 | 18.93 | 22.76* | 22.43* | 22.23† |

**(e) BiLSTM+CRF**

| | | AB | FT | IN | RW | MD | ER | DC | WS | SS | CS |
|---|---|---|---|---|---|---|---|---|---|---|---|
| PMC-1316 | F1@3 | 10.63 | 10.8 | 11.33† | 9.74† | 7.48 | 8.7 | 10.57† | 11.63 | 13.56† | 13.11 |
| | F1@5 | 10.28 | 10.3 | 11.57† | 11.36† | 6.87 | 8.13 | 9.42† | 13.07 | 12.03† | 13.82 |
| | F1@10 | 10.17 | 9.54 | 11.17† | 12.19† | 6.65 | 7.62 | 8.57† | 11.35 | 12.41† | 13.21 |
| LIS-2000 | F1@3 | 17.06 | 19.43 | 20.67† | 15.4† | 14.96† | 14.11 | 16.62† | 22.56* | 21.32† | 22.96 |
| | F1@5 | 17.12 | 20.82 | 20.27† | 17.05† | 14.58 | 13.07 | 17.14† | 19.31* | 21.01* | 23.03 |
| | F1@10 | 16.85 | 16.85 | 20.13† | 17.13† | 14.71 | 12.52 | 16.04† | 15.65† | 15.51† | 20.65 |
| IEEE-2000 | F1@3 | 17.81 | 19.75 | 20.51† | 13.68 | 16.01 | 13.38 | 16.31† | 21.14 | 19.75* | 22.84 |
| | F1@5 | 18.13 | 20.17 | 21.34† | 13.95 | 16.95† | 13.66 | 15.98† | 20.88 | 20.46† | 21.07 |
| | F1@10 | 18.92 | 16.7 | 20.39† | 13.81 | 15.3† | 12.81 | 15.93† | 18.04 | 17.96† | 19.12† |

**(f) BERT-BiLSTM-CRF**

| | | AB | FT | IN | RW | MD | ER | DC | WS | SS | CS |
|---|---|---|---|---|---|---|---|---|---|---|---|
| PMC-1316 | F1@3 | 23.93 | 16.49 | 26.27† | 21.81† | 22.16† | 20.9† | 21.08 | 22.91† | 23.69*† | 25.99 |
| | F1@5 | 24.83 | 18.89 | 22.76† | 21.89† | 20.55† | 22.68† | 22.65† | 23.13† | 27.01*† | 27.69 |
| | F1@10 | 24.66 | 16.62 | 22.55† | 21.87† | 20.76† | 19.94† | 22.13† | 23.47† | 25.46† | 26.73 |
| LIS-2000 | F1@3 | 22.54 | 19.62 | 20.92† | 20.92† | 19.4 | 21.43† | 20.62 | 25.31† | 24.63† | 27.67 |
| | F1@5 | 24.36 | 19.6 | 22.63† | 22.63† | 19.78 | 19.94 | 19.6 | 24.41† | 24.59† | 25.29 |
| | F1@10 | 23.02 | 18.55 | 22.5† | 22.5† | 17.19 | 17.1 | 18.55 | 23.63† | 22.86† | 22.9† |
| IEEE-2000 | F1@3 | 26.36 | 17.46 | 21.23† | 20.63† | 18.21 | 19.12 | 20.09† | 27.22† | 28.77*† | 28.07 |
| | F1@5 | 29.27 | 19.75 | 21.84† | 18.83† | 17.4 | 18.16 | 21.85† | 28.3† | 29.81† | 28.06 |
| | F1@10 | 29.71 | 17.73 | 21.15† | 18.15† | 17.17 | 16.3 | 21.04† | 22.99† | 23.22† | 24.29 |

**Fig. 5. Performance of KPE models on *Corpus-MH***

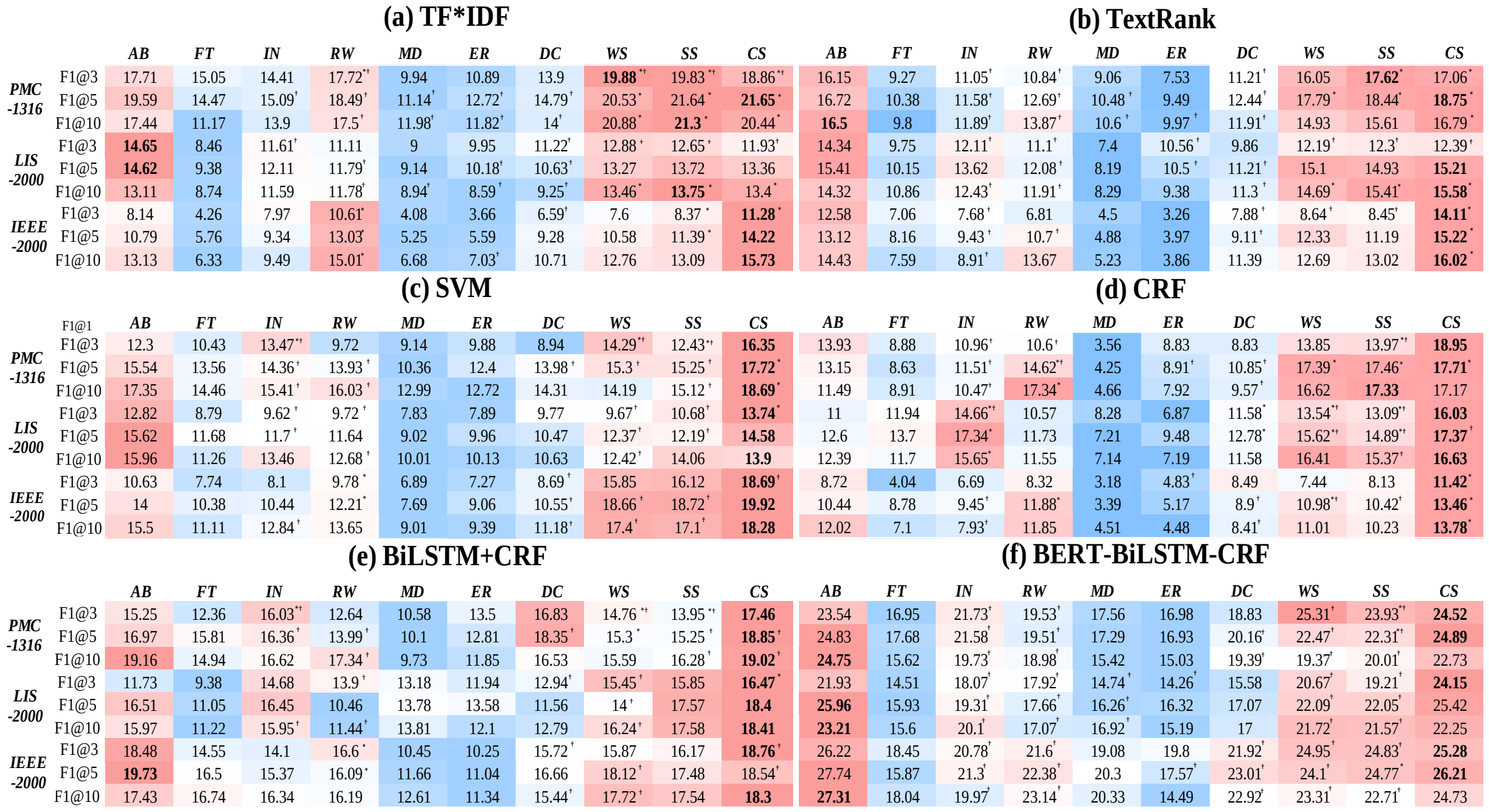

**Fig. 6. Performance of KPE models on *Corpus-ML***

**Table 12.** KPE performance with different methods.

| Models | Methods | PMC | | | LIS | | | IEEE | | |
|---|---|---|---|---|---|---|---|---|---|---|
| | | $F_1$@3 | $F_1$@5 | $F_1$@10 | $F_1$@3 | $F_1$@5 | $F_1$@10 | $F_1$@3 | $F_1$@5 | $F_1$@10 |
| TF*IDF | Nguyen | 15 | 16.43 | 14.26 | 9.85 | 12.76 | 11.16 | 6.20 | 6.69 | 6.16 |
| | Kontoulis | 15.24 | 17.2 | 15.12 | 10.96 | 10.94 | 10.76 | 4.79 | 4.89 | 5.81 |
| | Zhang | **18.48** | **19.3** | 16.7 | **15.41** | 16.43 | 13.08 | **10.68** | **11.52** | 10.97 |
| | CS | 14.3 | 15.77 | **17.62** | 15.29 | **18.85** | **19.23** | 8.87 | 10.12 | **12.5** |
| TextRank | Nguyen | 11.76 | 12.99 | 12.81 | 14.16 | 11.97 | 11.04 | 6.16 | 8.41 | 7.32 |
| | Kontoulis | 10.05 | 11.23 | 11.51 | 9.41 | 9.13 | 9.27 | 2.96 | 3.66 | 4.59 |
| | Zhang | 12.85 | 15.65 | 15.28 | 15.03 | 15.65 | 12.8 | **11.94** | **12.56** | 10.28 |
| | CS | **12.94** | **15.69** | **15.38** | **16.18** | **18.75** | **19.25** | 9.99 | 10.42 | **11.34** |
| SVM | Nguyen | 10.22 | 16.99 | **17.52** | 16.79 | 19.22 | 18.27 | 10.68 | 12.44 | 12.75 |
| | Kontoulis | 12.11 | 15.02 | 16.96 | 14.57 | 17.24 | 15.11 | 12.04 | 13.55 | 11.78 |
| | Zhang | **19.3** | 17.71 | 17.41 | **21.82** | **21.19** | **21.48** | **19.15** | **19.96** | **17.39** |
| | CS | 17.46 | **18.48** | 17.02 | 17.41 | 20.56 | 19.42 | 17.48 | 18.16 | 16.98 |
| CRF | Nguyen | 14.94 | 17.99 | 19.79 | 18.28 | 20.91 | 20.76 | 16.7 | 18.26 | 19.57 |
| | Kontoulis | 14.25 | 14.68 | 14.68 | 13.41 | 14.34 | 14.62 | 14.36 | 14.72 | 14.65 |
| | Zhang | 18.2 | 19.06 | **20.63** | 20.7 | 22.85 | **23.26** | 19.9 | 21.01 | 21.32 |
| | CS | **19.13** | **20.01** | 19.36 | **21.98** | **23.01** | 21.77 | **23.56** | **24.05** | **21.68** |
| BiLSTM-CRF | Nguyen | 4.90 | 6.37 | 8.80 | 7.67 | 9.40 | 11.84 | 4.04 | 6.95 | 12.06 |
| | Kontoulis | 6.08 | 6.83 | 6.91 | 11.33 | 11.48 | 11.48 | 9.49 | 9.92 | 9.87 |
| | Zhang | 6.72 | 8.82 | 9.49 | 12.45 | 14.60 | 17.00 | 9.76 | 11.91 | 12.94 |
| | CS | **12.76** | **12.85** | **11.94** | **23.63** | **23.08** | **19.28** | **19.93** | **17.79** | **16.19** |
| BERT-BiLSTM-CRF | Nguyen | 16.58 | 18.61 | 18.6 | 18.52 | 19.48 | 17.11 | 23.77 | 24.46 | 20.34 |
| | Kontoulis | 19.12 | 22.24 | 21.78 | 15.90 | 18.93 | 19.29 | 15.94 | 17.58 | 17.27 |
| | Zhang | 23.34 | 24.22 | 20.49 | 23.35 | 23.97 | 22.31 | 25.31 | 24.41 | 19.03 |
| | CS | **26.34** | **28.13** | **25.14** | **28.94** | **27.55** | **24.15** | **26.68** | **26.84** | **22.04** |

**Note:** The bold font represents the method that achieves the best performance on a metric.

To access the KPE results with existing studies, we replicated the studies by Nguyen & Luong (2010), Kontoulis et al. (2021), and Zhang et al. (2022) using *Corpus-PH*. The results, denoted as *Nguyen*, *Kontoulis*, and *Zhang*, are presented in Table 12. The results show that *CS* outperforms those of *Nguyen* and *Kontoulis*. *CS* outperforms *Zhang*'s KPE results across all models except for *SVM*. Notably, the *BERT-BiLSTM-CRF* model achieves the highest performance, with an average improvement of 4.79%. Thus, this method effectively enhances KPE performance and outperforms previous approaches.

### 5.3 Impact of Section Structure Recognition Quality on KPE

To demonstrate the impact of section structure recognition quality on KPE performance, we calculated the differences between the integrated results (*CS*) and the baseline variables *AB* and *FT*, yielding two variables: *CS-AB* and *CS-FT*. These variables reveal the performance trends across three corpora (*Corpus-PH, Corpus-MH, and Corpus-ML*), clarifying the influence of section structure on KPE.

**Table 13.** Normality Test for Paired Sample Differences of CS-AB and CS-FT Across Different Corpora

| Data Item | Kolmogorov-Smirnov | | |
|---|---|---|---|
| | Statistic | Value of *Free* | Significance |
| $CS\text{-}AB_{Corpus\text{-}PH} - CS\text{-}AB_{Corpus\text{-}MH}$ | 0.124 | 54 | 0.038 |
| $CS\text{-}AB_{Corpus\text{-}PH} - CS\text{-}AB_{Corpus\text{-}ML}$ | 0.080 | 54 | 0.200 |
| $CS\text{-}AB_{Corpus\text{-}MH} - CS\text{-}AB_{Corpus\text{-}ML}$ | 0.107 | 54 | 0.184 |
| $CS\text{-}FT_{Corpus\text{-}PH} - CS\text{-}FT_{Corpus\text{-}MH}$ | 0.082 | 54 | 0.200 |
| $CS\text{-}FT_{Corpus\text{-}PH} - CS\text{-}FT_{Corpus\text{-}ML}$ | 0.185 | 54 | 0.000 |
| $CS\text{-}FT_{Corpus\text{-}MH} - CS\text{-}FT_{Corpus\text{-}ML}$ | 0.186 | 54 | 0.000 |

Paired-sample tests were conducted, and the normality of differences was evaluated using the Kolmogorov-Smirnov test. Table 13 shows that most differences are not normally distributed, except for specific cases (e.g., CS-AB between Corpus-PH and Corpus-ML). The Wilcoxon signed-rank test was used to identify significant differences between paired samples.

Table 14 shows the Wilcoxon signed-rank test results for paired samples of *CS-AB* and *CS-FT* across corpora with varying section structure quality. The results indicate no significant difference in *CS-AB* performance between *Corpus-PH* and *Corpus-MH* but reveal significant differences between *Corpus-PH* and *Corpus-ML*, as well as *Corpus-MH* and *Corpus-ML*. This suggests that section structure recognition quality affects KPE performance. Specifically, there is no difference in performance improvements between high- and medium-quality section structure annotations and using "Title + Abstract" (*AB*) for KPE.

**Table 14.** Wilcoxon Signed-Rank Test for Paired Samples of CS-AB and CS-FT Across Different Corpora

| Data Item | Z | Significance |
|---|---|---|
| $Corpus\text{-}PH_{CS\text{-}AB} - Corpus\text{-}MH_{CS\text{-}AB}$ | $-1.395^{a}$ | 0.163 |
| $Corpus\text{-}PH_{CS\text{-}AB} - Corpus\text{-}ML_{CS\text{-}AB}$ | $-4.202^{b}$ | $0.000^{*}$ |
| $Corpus\text{-}MH_{CS\text{-}AB} - Corpus\text{-}ML_{CS\text{-}AB}$ | $-3.754^{b}$ | $0.000^{*}$ |
| $Corpus\text{-}PH_{CS\text{-}FT} - Corpus\text{-}MH_{CS\text{-}FT}$ | $-1.537^{a}$ | 0.124 |
| $Corpus\text{-}PH_{CS\text{-}FT} - Corpus\text{-}ML_{CS\text{-}FT}$ | $-.809^{b}$ | 0.418 |
| $Corpus\text{-}MH_{CS\text{-}FT} - Corpus\text{-}ML_{CS\text{-}FT}$ | $-2.256^{b}$ | $0.024^{*}$ |

**Note**： *:Indicates statistical significance at $\alpha = 0.05$; [a]:Based on positive ranks; [b]:Based on negative ranks.

However, significant differences are observed when comparing high- and low-quality as well as medium- and low-quality section structure annotations. Similar trends are partially observed in *CS-FT*. The findings indicate that while the quality of section structure recognition affects keyword extraction performance, it does not imply that higher-quality annotations always yield the best performance or that lower-quality annotations consistently yield the worst performance. For example, as shown in Fig. 6, *CS* results based on section structure using the TextRank algorithm still achieve optimal performance on the *PMC* and *IEEE* datasets, even when derived from low-quality annotations in *Corpus-ML*.

### 5.4 Case study

In this section, we present two case studies employing the *BERT-BILSTM+CRF* model, renowned for the best performance. The first one assesses the impact of different academic sections on KPE. The second validates the efficacy of the proposed section-based keyphrase integration algorithm, which is proposed in this paper.

#### 5.4.1 Case study 1: An Example of KPE Based on Different Section Texts

**Table 15.** Performance of KPE models on Corpus-ML

**Title:** Loss of Productivity Due to Neck/Shoulder Symptoms and Hand/Arm Symptoms: Results from the PROMO-Study

**Author Keyphrases:** Productivity, Musculoskeletal symptoms, Presenteeism, Computer workers, Psychosocial factors

**Predicted Keyphrases (N = 10):**

| | |
|---|---|
| **AB** | product loss, sick absenc, physic activ, comput worker, product, job |
| **FT** | physic activ, person trait, problem, sick leav, insur compani, comput, comput worker, job satisfact, psychosoci, use |
| **IN** | sick absenc, product loss, occup disabl, chronic pain, chronic occup disabl, chronic disabl, comput worker, offic worker, reintegr, product |
| **RW** | comput, physic activ, comput worker, comput worker product, comput usag, contract, product loss, product, hand arm, pain |
| **MD** | person, psychosoci load, psychosoci, product loss, reward, job satisfact, comput, product, satisfact |
| **ER** | sick absenc, comput use, product loss, mous posit, ergonom, job perform, comput worker, sick leav, comput worker product, physic activ |
| **DC** | sick absenc, absente, presente, product loss, product measur, product, comput, comput usag, psychosoci, chronic |

**Note**: The phrases highlighted in yellow is align with the ground truth, while blue highlights indicate predicted phrases that are synonymous or closely related to the ground truth.

The impact on KPE confirmed above may be questioned as not being brought about by the inclusion of section structure information, may also be attributed to the integration algorithm itself. In order to visually confirm the effect of section structure on keyphrase extraction, we give out a case study in Table 15. The distribution of highlights demonstrates excellent prediction results when combining different section texts, providing relevant contextual information to facilitate reader understanding. Specialized terms like "Musculoskeletal Symptom" may be challenging, but keyphrases such as "hand arm" and "chronic pain" offer more intuitive expressions. Using section-based keyphrases enhances accessibility and helps readers better grasp the article's meaning. After stemming, the keyphrases derived from section text align more closely with the ground truth. Thus, leveraging section text is beneficial for high-quality KPE and enhances readers' comprehension of the article.

### 5.4.2 Case study 2: An Example of KPE Based on the Integrating Algorithm

To assess the effectiveness of our proposed section-based integration method on KPE model, we compare the prediction outcomes for *AB* and *FT* with the integrated section result denoted as *CS*. As shown in Table 16, we randomly select one research paper in *Corpus-PH* from each of the three domains.

**Table 16. Results of KPE from three domain with different method**

| Type | Paper1 | Paper2 | Paper3 |
|---|---|---|---|
| Domain | Biomedical science | Library and information science | Computer science |
| Title | Incidence and visual outcome of endophthalmitis associated with intraocular foreign bodies | Continuous usage of social networking sites: The effect of innovation and gratification attributes | Unifying Visual Attribute Learning with Object Recognition in a Multiplicative Framework |
| Author keywords | endophthalmitis(70), trauma(30), foreign bodies(8), visual acuity(4), risk factors(3) | social networking sites(5), continuous usage(4), innovation diffusion(26), internet behaviour(0), uses and gratifications(10) | attribute learning(29), zero-shot learning(17), image understanding(0) |
| AB | cultures, **trauma**, pars plana vitrectomy, visual outcome, penetrating eye trauma | **social networking sites**, user motivation, social norms, innovation, facebook | deep learning, **zero-shot learning**, **attribute learning**, typical learning, attributes |
| FT | cryotherapy, penetrating ocular trauma, cultures, encapsulation, referral | mass media, intrinsic motivations, psychological motivations, continuous intention, innovative technologies | object categorization, object detection, weight decay, semantic embedding, **zero-shot learning** |
| CS | **endophthalmitis**, intraocular foreign bodies, vitrectomies, **trauma**, visual outcome | innovation, **innovation diffusion**, play, **social networking sites**, social network | attribute prediction, **attribute learning**, object recognition, visual attribute learning, deep learning |

According to Table 16, keyphrases extracted from abstracts (AB) exhibit significantly better quality than those from full text (FT). Specifically, in Paper1 and Paper 2, keyphrases integrated in section-level (CS) demonstrate superior performance. However, for Paper3, the highest accuracy is achieved using AB section text. For instance, in Paper1, the term "endophthalmitis" frequently appears across various sections, facilitating its extraction through structural content. This illustrates how a KPE model utilizing structural content and an integration algorithm effectively captures scattered author keywords. Additionally, the study found that some predicted keyphrases, such as "object recognition" and "visual attribute learning" from CS, serve as effective descriptors even if they are not in the author's keyword list, as seen in Paper3.

## 6. Discussion

Compared to titles and abstracts, author keywords offer broader coverage within the full text of academic articles

(Zhang et al., 2022). This suggests that utilizing the full text of academic articles may enhance the performance of KPE. Moreover, with the growing momentum of the open access movement, accessing the full text of academic articles has become increasingly convenient, thus further promoting the utilization of full text for enhancing KPE performance. However, the full text of academic articles also contains a great number of irrelevant phrases, which can hinder the accuracy of KPE. To address this challenge, this paper employs a section structure classification method to acquire section structure information from academic articles, intending to leverage structural features and section texts to enhance the performance of KPE.

## 6.1 Implication

This paper presents the implications of the study in terms of both theoretical significance and practical application.

### 6.1.1 Theoretical implication

Firstly, this paper introduces an interdisciplinary, section-annotated KPE corpus, providing both data and theoretical support for research in other NLP tasks and languages. Furthermore, we establishes a standard of section classification utilizing machine learning and deep learning technologies, which serves as a theoretical reference for subsequent section-related NLP research. In addition, this study offers valuable insight for research on knowledge entity extraction through full-text analysis. As keyphrases serve as knowledge entities, the section-aware KPE technique can be applied to other tasks as well. Moreover, this paper proposes an enhanced keyphrase extraction framework that comprehensively considers both section structure features and textual content, enriching the theoretical foundation of the keyword extraction task. The results validate the potential value of section structure information not only for KPE but also for other knowledge entity extraction tasks. Academic articles typically exhibit a distinct sectional organization, with varying significance attributed to each section. While Zhou & Li (2020) argued for the benefits of identifying sections in academic articles for filtering noise during entity and relationship extraction, their claims lacked experimental validation. The experiments in this study uncover the potential value of section structure, thereby substantiating their views.

### 6.1.2 Practical implication

In recent years, the exponential increase in the volume of academic publications has made it challenging for researchers to swiftly access relevant literature. This highlights the need for more accurate document retrieval. Our method can provide more representative and accurate keyphrases, thus providing more relative and easy-to-understand search results for readers. Secondly, KPE based on section structure offers keyphrases that encompass more contextual information. These phrases assist readers in better understanding the essential concepts and focal areas of an article. Additionally, individuals involved in abstract writing or literature reviews can benefit from familiarity with section structure, as it enables them to extract and summarize critical information accurately, effectively conveying the main theme of the article. Moreover, this study can support researchers in conducting topic analyses within specific research domains. Keyphrases serve to depict the research themes of an academic article. Some journals or conference papers, such as PLOS ONE and ACL, may lack keyphrases. Consequently, our method can be used to assign keyphrases to articles missing author keywords, thereby enriching the articles' keyphrase content.

### 6.2 Limitation

This study also presents limitations that can be addressed in future work. First, the list of matching words used for automatic section classification in this paper is relatively simple, potentially overlooking certain phrases crucial for determining section categories, which may result in classification errors. Second, the section structure classification framework proposed in this paper was developed based on the characteristics of the experimental datasets, utilizing a coarse-grained approach solely reliant on section annotations. In reality, more intricate frameworks can identify the sections based on the analysis on sentence and clause levels, respectively. Implementing a more fine-grained structure classification framework for KPE from academic articles may yield improved performance. Third, this paper only considered using structural features and section texts from the section structure information for KPE, without a consensus on the optimal utilization of section structure information. Lastly, the paper treated phrases from different section texts as equally important. By weighting phrases from distinct section texts and re-evaluating each phrase's importance, further insights may be uncovered.

## 7. Conclusion and future works

The objective of this work is to demonstrate that section provides richer contextual and structural information than individual sentences, thereby enabling more accurate capture of keyphrases. Specifically, the innovation of this paper lies in three aspects: corpus construction, section classification, and experimental design. Firstly, we selected diverse realm of academic papers, enhancing the experiment's adaptability across various domains. Secondly, it proposed a standard for section classification and categorizes the corpus into clear, moderately clear, and ambiguous section structures, thereby providing high-quality data support for subsequent section-based KPE experiments. Notably, the paper discusses from the dimensions of section structure features and section text how section information affects the performance of KPE on corpora of different section clarity. The result demonstrates that section information positively impacts KPE performance. Furthermore, texts with clearer section structures generally exhibit better KPE performance; however, the performance of KPE on texts with ambiguous section structures is not necessarily worse, possibly due to other inherent text features, warranting further investigation in subsequent research.

In conclusion, this paper highlights the importance of incorporating section structure information into KPE. This methodology effectively addresses the challenges posed by noisy phrases and lengthy sequences within the full text of academic articles. Additionally, considering the limitations acknowledged in this paper, future work will investigate the impact of different types of structure classification frameworks on KPE and investigate improved methods for leveraging section structure information in academic articles to enhance KPE performance. Furthermore, the study's findings revealed that some phrases, although not similar to the author-annotated keywords, serve as significant descriptors crucial for understanding key information. These phrases might enrich the semantic context of the keywords, aiding readers in better comprehending the essence of the article.

## Endnotes

1. https://pmc.ncbi.nlm.nih.gov/about/intro/

## Acknowledgments

This study is supported by the National Natural Science Foundation of China (Grant No. 72074113) and Jiangsu Province Graduate Student Research Practice Innovation Program (Grant No. KYCX23_0609).

## Declarations

The author(s) declared no potential conflicts of interest with respect to the research, authorship, and/or publication of this article.